# Predicting Mergers and Acquisitions using Graph-based Deep Learning

Keenan Venuti, Northwestern University

**Abstract**
The graph data structure is a staple in mathematics, yet graph-based machine learning is a relatively green field within the domain of data science. Recent advances in graph-based ML and open source implementations of relevant algorithms are allowing researchers to apply methods created in academia to real-world datasets. The goal of this project was to utilize a popular graph machine learning framework, GraphSAGE, to predict mergers and acquisitions (M&A) of enterprise companies. The results were promising, as the model predicted with 81.79% accuracy on a validation dataset. Given the abundance of data sources and algorithmic decision making within financial data science, graph-based machine learning offers a performant, yet non-traditional approach to generating *alpha*.

**Introduction**
The intersection between data science and financial markets has a robust historical foundation and continues to evolve every day. Automated algorithmic trading, financial fraud detection, and real-time news sentiment analysis are examples of applications of modern machine learning within the world of finance. Given the accessibility of powerful computers, financial data, and programmers able to capitalize on future trends, finding signals within a market requires strong mathematical theory, an unorthodox discovery approach, and luck. Once a signal has been discovered, sometimes referred to as *alpha*, huge dividends can be yielded given a large enough investment. This competitive edge may require taking a new approach to finding trends within a well researched domain, such as mergers and acquisitions. Graph-based machine learning offers new methodologies to interpret and make predictions on massive graphs, using entity to entity relationships.

*Graph Data*
The graph is a historical mathematical concept that is core to frameworks like Markov Decision Processes, Page Rank web search, and social networks. Graphs are defined as a set of nodes and edges representing relationships between entities. Nodes, or entities, can have distinct or shared properties, but are never identical within a graph. For example, there won't be two nodes that represent the same company entity. Undirected edges are the links that unite nodes based on a shared, unidirectional relationship. Edges can also be directed to reflect when a relationship is not mutual. Graphs as a data structure contain different information to traditional vectors, matrices, and tensors. This is because graphs exist in a non-euclidean space; a node's position is entirely determined by its neighbors. These relationships among neighboring nodes is core to analyzing graphs and performing graph-based machine learning.

**Literature Review**
*Predicting Mergers and Acquisitions*
Using machine learning to predict mergers and acquisitions is not a new sub-domain of data science or financial research. Researchers analyzing a Greek enterprise company dataset created several random forest models to predict target acquisitions (Adelaja, Nayga Jr, and Farooq 1999, 15). Another set of researchers experimented with natural language processing to analyze company fillings, and predict acquisitions and targets (Moriarty, Ly, Lan, and McIntosh 2019, 1). While these research projects studied separate domains of M&A, their approaches were fairly similar. Most research groups found a dataset on profits and losses and implemented traditional machine learning techniques like logistic regression and random forests. Their results are promising, but limiting as each study requires its own specific financial dataset. For example, if one company in the dataset hadn't disclosed their financial performance, that datum would have to be ignored. Further, the datasets also fail to capture signals between companies. If company X owns company Y, and company Z is a direct competitor to Y, it is highly possible that X may acquire Z in an attempt to reduce competition.

*Graphs to Solve Problems*
Graph data structures offer flexibility in capturing the relationships between entities (via heterogeneous edge types), and are highly tolerant of missing or corrupt data. Further, the creation of a financial enterprise graph database would allow for future work outside of M&A prediction. Traditionally, graphs and graph databases have been used to represent communities like computer networks, supply chain distribution, and social networks (Announcing: Graph-Native Machine Learning in Neo4j! 2020, 1). Given the relationship-heavy nature of these data sources, any other methodology of storage would not be appropriate. Graph databases are not limited to these domains, as they can be used for any dataset dependent on relationships among entities. However, because of the non-euclidean nature of graphs, leveraging them for analytics and machine learning is more challenging than using plug and play algorithms.

*Graph Machine Learning Algorithms*

To fully leverage graph data for analytics and machine learning, some transformations are made to encode nodes and edges into euclidean spaces. Graphs can be quantified using adjacency matrices. They are an NxN matrix where N is the count of nodes in a graph and each row, i,j, represents a relationship between node i and node j. They are limited to homogeneous directed edge types and an extremely dimensional form of data. To make graph data more useful, it is standard to use an embedding algorithm (Knyazev 2020, 1). Embedding algorithms reduce the dimensionality of a dataset by capturing the relationships between individual entities. Two of the most popular embedding algorithms are Node2Vec and DeepWalk (Zhang, Li, Xia, Wang, and Jin 2020, 1). Essentially, the algorithms simulate random walks to construct a vector representation of each node based on their neighbors. Embedding algorithms are not only used for nodes as they can also be used to embed subgraphs or edges (Zhang, Li, Xia, Wang, and Jin 2020, 2). All share the same principle, to vectorize the graph. Once represented as a vector, the data are ready to be used for machine learning.

Within the last 5 years, there have been several new graph neural network models published within academia (Grattarola and Alippi 2020, 2). A graph neural network can be surmised by its two core components; a message passing layer for non-linear transformations of nodes to vectors and a graph pooling layer to reduce the size of the messages passed. The core algorithm for most graph-based deep learning is as follows:
- Initialize a vector for every node in a graph (could be based on node type, properties, or randomly)
- Establish a downstream task such as node classification or link prediction
- **Message Passing**: Create a specialized layer, where every node receives all of its neighbors' weights
    - (alternatively, create n layers to receive weights from nodes that are a walk of distance n from the current node)
- Apply a transformation and aggregation function on the adjacent nodes' weights
    - The transformation function's weights are updated every time the network is backwards propagated
- **Graph Pooling**: Use the results of the previous step to update the current node's weights
    - This update is performed every time the network is forward propagated
- With the nodes' weights being trained and updated, the output is a matrix of size NxM for N nodes in a graph each embedded into a vector of size M
- Use the matrix for the downstream machine learning task

(Allamanis 2020)

Variants exist to this methodology, but most use the same message passing algorithm (Ma 2020, 1). For this project, I used a popular variant named GraphSAGE. The GraphSAGE model was developed by Stanford University's SNAP research group (Hamilton, Ying, and Leskovec 2017, 2). The model is defined as a, "framework for inductive representation learning on large graphs" (Leskovec 2020). GraphSAGE is particularly adept at generating lower dimensional embeddings for graphs with important node attributes. Most graph embedding models are transductive and unable to perform embeddings for new graphs. GraphSAGE is considered inductive as its message passing methodology allows for predictions with previously unseen graphs (Hamilton, Ying, and Leskovec 2017, 2).

**Methods**
*Data Acquisition*
Finding a data source on enterprise companies, and their relationship without sponsorship from business intelligence organizations was a challenge. While datasets of financial performance exist, they fail to capture the entity to entity relationships required for this project. Ultimately, I found a fairly comprehensive open source dataset to pull data from. Wikidata is one of the world's largest triplestores, holding ~1.1 billion triples. Triplestores hold semantic triples which are essentially the relationships between two nodes. This is very similar to data stored in graph databases (like Neo4j) without the aggregation step of combining identical nodes. To pull from Wikidata, I developed two SPARQL queries (Wikidata's query engine). One extracted all companies within the top 30 countries (ranked by GDP) and the other extracted all companies that were owned by another company within the first query. To load the raw data into the popular graph database Neo4j, I had to convert the subject-predicate-object triples into nodes and edges. I used Python's popular Pandas library and a transformation script to ingest the data source. This resulted in 61,026 nodes with 115,027 connections, 2,985 of which were ownerships, the target prediction edge.

*Data Exploration*
Data exploration typically requires summary statistics of a dataset's features, measuring the response variable's distribution, and identifying anomalies like outliers and extreme values via statistical measures of significance. While data exploration with graph datasets follows the same core principles, the methodology differs. Graph data is more similar to language data; unstructured, but containing many latent rules and trends. Where an NLP researcher may use metrics like word counts or average sentence length to understand their dataset, graph analytics rely heavily on traditional graph mathematics. This includes metrics like clustering coefficients, average outgoing/incoming edge count, and count of triangles. Metrics like these allow the researcher to evaluate the connectedness of each node, and general fluidity or sparseness of the graph. Graphs often follow a power law distribution for outgoing edge counts, as there are frequently a few highly connected nodes and many sparsely connected nodes. This dataset followed a similar edge count distribution:

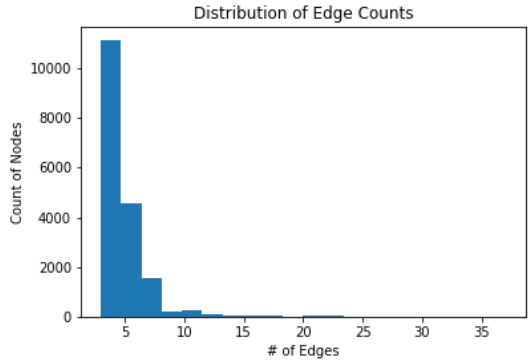

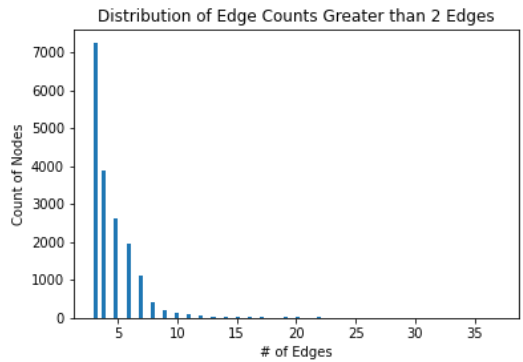

Clustering coefficients measure the tendency of nodes in a graph to cluster together. As shown below, the vast majority of nodes in the graph have a low clustering coefficient of 0. However, there are some nodes that are highly clustered:

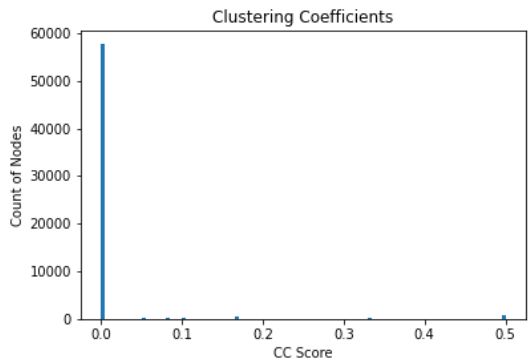

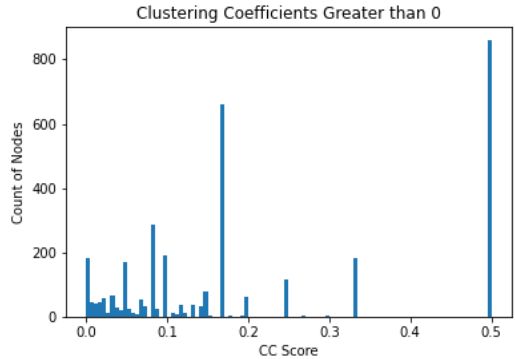

*ML Data Preparation*
Where there exists machine learning libraries like Sklearn, TensorFlow, and H2O for out-of-the-box tabular machine learning, applying similar algorithms to graph data is more challenging. Machine learning for large graphs is a relatively new concept and not supported by major machine learning frameworks. Fortunately, there is an open source library, StellarGraph that is equipped to perform train/test splitting, model implementation, and even GPU based training.

For this project, I leveraged the StellarGraph Heterogeneous GraphSAGE link prediction model. To prepare the data, I used StellarGraph's train test splitting functionality. The graph train test splitting algorithm consists of the following steps:
- Randomly drop a set of target edges (edges the model will predict), save them for testing, and negative sample non-existent edges to evaluate the model with via binary cross entropy
- Using the graph with dropped edges, again drop a random set of target edges, save them for training, and negative sample non-existent edges to train the model
- Any remaining target edges in the doubly dropped graph will become features in link prediction

The resulting double dropped graph may still contain target edges. In graph, a target edge can also be considered a feature as a relationship between two nodes and may help in edge prediction for a third node. Ultimately, the researcher must decide to use all target edges in the training dataset, or leave a portion within the sampled graph, to be used as features. In this project, I chose the former option, opting to use all target edges for training and testing. Finally, I randomly initialized node embeddings to prevent data leakage.

*Graph ML Implementation*
The final model was a GraphSAGE Heterogeneous model with 28,704 trainable parameters. Below is a

diagram of the model:

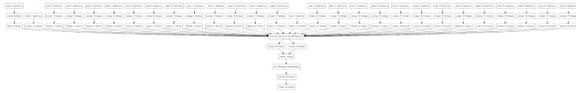

It is composed of 25 incoming layers; each incoming layer represents a node's neighbor's weights to be used in training and evaluation during forward propagation. It is set to 25, as the training data's nodes had a maximum of 25 incoming edges. Each incoming layer then goes through a dropout layer, with drop probability set to .6. Between the incoming layers and outgoing layer exists a Relu activation function layer of density 32. The final layer is a simple reshape to create the binary classification.

**Results**
*Model Selection and Performance*

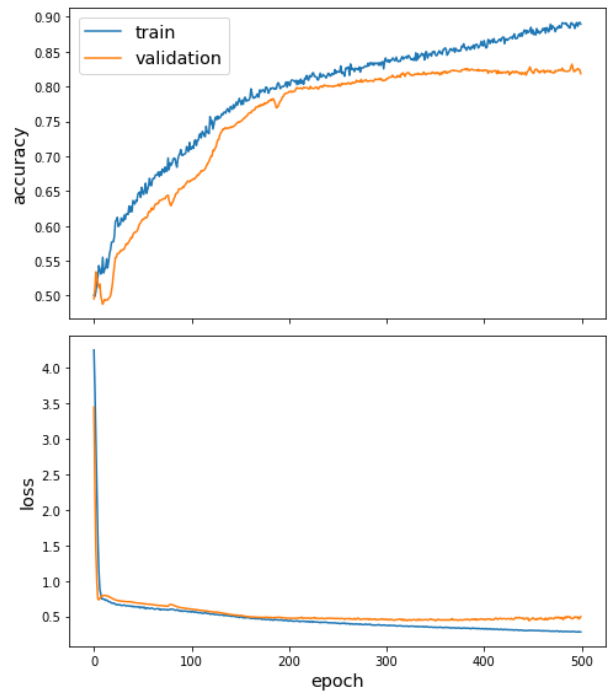

The best performing model, described above, had a training accuracy of 88.58% and a validation accuracy of 81.79%. Loss was calculated with binary cross-entropy; the training set had a loss of .2530 and validation set had .5078. Adjacent is a visualization of it's training and validation accuracy and loss performance as it was trained. The closeness of the two curves shows a healthy train, without much over fitting. Consistent with all the models I trained, towards epoch 400, the curves start to diverge implying the model began over-fitting to noise in the training data rather than signal.

**Analysis and Interpretation**
*Hyperparameter Tuning*
While the final model had a fairly straightforward implementation, it required some testing to find an optimal set of hyper-parameters. The most important hyper-parameters for the GraphSAGE algorithm are:

| Number of samples | Number of layers/iterations in the model |
| --- | --- |
| Dropout | Probability of ignoring a hidden layer node during training |
| Hidden Layer Size | Number and size of hidden layers within the activation function layer |
| Target edge features | Number of target edges to use as features |
| Embedding Method | Use concatenated node embeddings to generate link embeddings |

When experimenting, I found that some of the model variants suffered from three typical machine learning problems; below are the problems and their solutions:

| Model was slow to train/build performance | Increased learning rate |
| --- | --- |
| Model had high bias/underfitting | Increased model complexity |
| Model had high variance/overfitting | Increased dropout/randomly initialized embeddings |

*Final Variant Choice*
In graph datasets, target edges can be labels or used as features. Sometimes, it would be helpful to know that company X acquired company Y when evaluating the probability of company X acquiring company Z. Here, knowledge of X acquiring Y would be a feature and could not be used as a prediction label. Graph models give the choice of how many target edges become features instead of testing/training labels. Ultimately, after trial and error, I found that including target edges didn't significantly increase performance. To eliminate all potential data leakage, I excluded target edges as features.

**Conclusions**
The GraphSAGE model out-performed comparable models designed to predict M&A. A group of researchers using a Greek enterprise company dataset, created a model with 47.62% target and 83% non-target classification performance (Athanasios, Georgopoulos, and Siriopoulos 2007, 10). One research team focused specifically on predicting M&A in the food industry had a better accuracy score of 85.7% (Adelaja, Nayga Jr, and Farooq 1999, 15). Given the comparable performance, graph-based models show promising utility when analyzing financial data. Any relationship heavy data source like enterprise company relationships will work well for these types of analyses. The results confirmed the initial hypothesis that a non-traditional method for predicting M&A has real-world utility. I also found that these types of models help identify which edge types and node properties are most effective in link prediction. Graph models supplemented with other model types yield interesting futures in corporate business intelligence.

**Future Work**
*Temporal Based GNNs*
One important improvement to the model used in this project would be the introduction of time-series data into the graph dataset. Like most relationships, mergers and acquisitions are not static. They form over time, edges that didn't exist at time step t may exist at time step t+1. This principle is true for all relationships within my graph dataset. As a result, applying a static machine learning algorithm only allows prediction of current relationships. This greatly reduces the real-world applicability of the model. Excluding time steps from the data was a decision based on the availability of time based edges in the graph dataset. In future work, I'd like to re-implement the dataset with time based edges and fit LSTM based GNNs to the graph dataset and evaluate link prediction performance at each time step. Within temporal GNNs, there exists several variants (Chen, Xu, Wu, and Zheng, 2018, 1) (Chen, Zhang, Xu, Fu, Zhang, and Xuan 2019, 1) (Mutinda, Nakashima, Takeuchi, Sasaki, and Onizuka 2019, 1).

*Supplemental NLP Triple Extraction*
Given the popularity of modern financial research, there is an abundance of data on enterprise company relationships. However, much of this data is held in unstructured text documents, and inaccessible through traditional machine learning. A news article that communicates that company X is a direct competitor with company Y would be instrumental in the formation of the graph database described in this paper. To extract this kind of triple, a modeler would need to implement a triple extraction algorithm, and apply it to terabytes of textual news data. This process was outside of the scope of this project, but would yield substantial increases in both the results and applicability of the work.